# Using 3D reconstruction from image motion to predict total leaf area in dwarf tomato plants

Dmitrii Usenko[1], David Helman[1,2], Chen Giladi[3,*]

1. Department of Soil and Water Sciences, Institute of Environmental Sciences, The Robert H. Smith Faculty of Agriculture, Food and Environment, The Hebrew University of Jerusalem, Rehovot 7610001, Israel
2. The Advanced School for Environmental Studies, The Hebrew University of Jerusalem, Jerusalem 9190401, Israel
3. Department of Mechanical Engineering, Sami Shamoon College of Engineering, 84 Jabotinsky St., Ashdod 77245, Israel

*Corresponding author: chengi1@sce.ac.il

Abstract

Accurate estimation of total leaf area (TLA) is essential for assessing plant growth, photosynthetic activity, and transpiration but remains a challenge for bushy plants like dwarf tomatoes. Traditional destructive methods and imaging-based techniques often fall short due to labor intensity, plant damage, or the inability to capture complex canopies. This study evaluated a non-destructive method combining sequential 3D reconstructions from RGB images and machine learning to estimate TLA for three dwarf tomato cultivars—*Mohamed*, *Hahms Gelbe Topftomate*, and *Red Robin*—grown under controlled greenhouse conditions. Two experiments, conducted in spring–summer and autumn–winter, included 73 plants, yielding 418 TLA measurements using an "onion" approach, where layers of leaves were sequentially removed and scanned. High-resolution videos were recorded from multiple angles for each plant, and 500 frames were extracted per plant for 3D reconstruction. Point clouds were created and processed, four reconstruction algorithms (*Alpha Shape*, *Marching Cubes*, *Poisson's*, and *Ball Pivoting*) were tested, and meshes were evaluated using seven regression models: Multivariable Linear Regression (*MLR*), Lasso Regression (*Lasso*), Ridge Regression (*Ridge-Reg*), Elastic Net Regression (*ENR*), Random Forest (*RF*), extreme gradient boosting (*XGBoost*), and Multilayer Perceptron (MLP). The *Alpha Shape* reconstruction ($\alpha = 3$) combined with *XGBoost* yielded the best performance, achieving an $R^2$ of 0.80 and MAE of 489 cm$^2$, with significant results across other model combinations. Results were lower when using data from different experiments as train and test datasets ($R^2 = 0.56$ and MAE = 579 cm$^2$). Feature importance analysis identified height, width, and surface area as the most predictive features. These findings demonstrate the robustness of our approach across variable environmental conditions and canopy structures. This scalable, automated TLA estimation method is particularly suited for urban farming and precision agriculture, offering practical implications for automated pruning, improved resource efficiency, and sustainable food production.

## 1. Introduction

Total leaf area (TLA) is a comprehensive metric describing the plant's growth and functioning. It is a primary metric that describes the plant's photosynthetic activity and transpiration capacity. Normalized by the plant's surface area, TLA may provide information on the canopy structure, which is crucial for understanding the plant's energy and resource efficiency. For example, reduced TLA is a sign of stress (Dong et al., 2019), while excessive biomass, indicated by a higher TLA, signifies lower water use efficiency (Glenn et al., 2006). Farmers often use pruning to reduce TLA in commercial crops to increase crop productivity (Budiarto et al., 2023). However, measuring and finding the optimum TLA of the crop are challenging tasks.

Traditional methods for measuring TLA can be broadly classified into destructive and non-destructive techniques. The most standard, widespread approach to measuring TLA is the destructive technique. It involves physically removing leaves from the plant and measuring their area using an area meter. Traditionally, leaves were placed over a grid of a known area. Nowadays, scanners can be used to measure leaf area more precisely. LiCOR 3100C leaf area meter is a widely used instrument that scans the leaves placed on a transparent conveyor belt and calculates the area based on light transmission. Flatbed scanners can also be used together with some applications of image analysis (Pandey & Singh, 2011). An obvious limitation of such an approach is that the plant must be destroyed to have its TLA measured, thus making it impossible to monitor leaf area dynamics.

Some portable scanners allow measuring the leaf's size while still attached to the plant (Karatassiou et al., 2015). However, this kind of sensor does not work well with bushy plants, such as dwarf tomato plants, simply because it is impossible to insert them into the canopy in a way that won't damage the plant. Also, measuring TLA is time-consuming and labor-intensive as the farmer must measure each leaf individually.

Numerous attempts have been made to develop a non-destructive technique for accurately estimating TLA. These methods leverage various imaging and computational approaches to overcome the limitations of traditional destructive techniques. Leaf size measurements (height and width) can be used to predict individual leaf areas with very high accuracy (e.g., Boyacı & Küçükönder (2022) for apples). Das et al. (2022) combined such an approach with imaging and machine learning (ML) to provide single-leaf areas.

When leaves are visible, some models try to predict the plant's TLA based on images. For example, Lee et al., 2022 used images of small plants with TLAs under 100 $cm^2$ to calculate their leaf areas using pixel count. Building on such a concept, Jiang et al. (2015) used multi-angle imagery to estimate TLA, capturing the plant's structure from different perspectives. They used a blue screen as a background to distinguish the plant pixels more easily. It was suggested that adding multiple image perspectives adds significantly more accuracy to the model than a single top-down image.

There were also successful attempts to estimate the leaf area of a plant using 3D reconstructions and point clouds. For example, Yang et al. (2024) presented a method of automating the 3D reconstruction of plants in a special setup using RGB cameras and image pre-processing pipelines. Feldman et al. (2021) used 3D point clouds to generate a plant's structure and analyze its projected leaf area. Gu et al. (2024) demonstrate a successful TLA estimation of a wheat plant using a specific multi-view imaging platform for plants (Wu et al., 2022). Thapa et al. (2018) created a similar setup using a rotational platform to generate accurate imagery and a LiDAR scanner to create a 3D point cloud. Wu et al. (2018) created a portable setup that allows the creation of high-quality LiDAR-generated point clouds. Reza et al. (2023) showed an application of an artificial neural network (ANN) to a set of plant images, yielding promising results in a particular range.

Though promising, these advanced, successful methods rely on carefully reconstructing each leaf that composes the plant's canopy. The chosen species and cultivars in the cited studies were suitable for that task since they usually had only a few overlapping leaves. However, many leaves may be "invisible" to the camera in a real-world case of more bushy plants. In such a scenario, which may be the case for many tomato plants, it would be impossible to reconstruct TLA from single-leaf images. Moreover, the abovementioned approaches require putting the plant on a special stand that would generate high-quality images, which takes time and requires intensive manual labor.

Here, we suggest taking the above approach one step further. We show that TLA can be assessed by creating a single 3D reconstruction of the entire plant from simple RGB images and combining it with observed TLA in machine learning algorithms. We chose three dwarf tomato plants for that purpose, as these represent the most difficult canopies for TLA assessment. Moreover, dwarf tomato plants, originally small recreational plants that produce fruit in a limited space, are suitable for cultivation in small urban spaces and may have potential for future commercial urban farming. We test 3D meshes of the plant's canopies reconstructed using different algorithms from the point clouds in five ML algorithms, seeking the best combination of mesh reconstruction and ML regression model.

## 2. Data & Methods

### 2.1. Plant material and growth conditions

#### 2.1.1. Tomato varieties

Dwarf tomato plants from the *Dwarf Tomato Project* (*Dwarf Tomato Project – A Co-Operative Venture*, n.d.) were used due to their growing popularity among amateur growers who may find our TLA estimation model useful (LeHoullier & Nuske-Small, n.d.). Three dwarf tomato cultivars were selected: *Mohamed*, *Hahms Gelbe Topftomate*, and *Red Robin*. These varieties were chosen due to their compact growth habits and suitability for urban farming applications, particularly in hydroponic systems where space efficiency is crucial.

All three dwarf tomato cultivars — *Mohamed*, *Hahms Gelbe Topftomate*, and *Red Robin* — share a compact, determinate growth habit suited for high-density planting and automated management systems. They produce cherry-sized tomatoes averaging 10–12 grams per fruit, with *Mohamed* and *Red Robin* yielding red fruits and *Hahms Gelbe Topftomate* producing bright yellow fruits, adding diversity to the dataset and potentially influencing light absorption. *Red Robin* is specifically known for its adaptability to confined spaces.

#### 2.1.2. Greenhouse conditions

The experiments were conducted in a controlled glasshouse at the Robert H. Smith Faculty of Agriculture, Food and Environment, The Hebrew University of Jerusalem, Rehovot, Israel (31.90445°N/34.80459°E; Jiang et al., 2022; Mulero et al., 2023). Two consecutive experiments were performed to assess the consistency and reliability of the method under varying seasonal conditions. The first experiment (Exp. #1) was run during the spring and summer when radiation is generally higher in this region. The second experiment (Exp. #2) was carried out in autumn and winter when radiation decreases.

Table 1 presents detailed information regarding the growing conditions of the two experiments.

While temperature and humidity were controlled to remain constant across both experiments, the primary environmental variable was the difference in natural solar radiation due to seasonal changes. Exp. #1 benefited from increased natural sunlight, promoting robust vegetative growth and potentially more significant leaf area. Exp. #2 experienced reduced natural sunlight due to the seasonal change. It led to a significant difference in yield between the two experiments, as the plants that received more radiation were more productive. However, in the scope of the current research, we regard it as a bonus opportunity to test and validate the models. These differences are valuable for testing the robustness of the 3D reconstruction and machine learning models under varying environmental conditions.

Table 1. Growing conditions in the two experiments (Exp. #1 and Exp. #2).

| Parameter | Exp. #1 | Exp. #2 |
| --- | --- | --- |
| *Timeline* | March 2023 — June 2023 (Spring to early summer) | September 2023 — December 2023 (Autumn to early winter) |
| *Temperature (fixed)* | 24 °C (daytime) / 18 °C (nighttime) | |
| *Relative humidity* | 70–75% (daytime) / 90–97% (nighttime) | |
| *Lighting conditions* | Natural sunlight. Average daily solar radiation was higher during this period, typical for Mediterranean spring and early summer. | Natural sunlight. Shorter daylight hours and lower natural solar radiation due to seasonal changes. |
| *Irrigation* | Plants were grown in 4 L pots filled with perlite, a soilless medium facilitating hydroponic growth. An automated drip irrigation system supplied 150 mL of water per plant five times daily, once in two hours, starting at 7:00 AM. | |
| *Nutrition* | Nutrient solutions were provided in the irrigation water, containing balanced macro and micronutrients suitable for tomato cultivation. | |
| *Pest control* | Integrated pest management strategies, including regular inspections and pesticide application, were employed to prevent infestations that could affect plant growth and leaf area. | |

Environmental parameters were continuously monitored in the experiments using a centralized climate control system equipped with temperature, humidity, and light intensity sensors. Sensors were connected to a central datalogger, and data were downloaded to monitor conditions, record any deviations for subsequent analysis, and adhere to the specified conditions.

An automated drip irrigation system delivered precise amounts of water to each plant. Each plant received 150 mL of water five times daily (at 7 a.m., 9 a.m., 11 a.m., 1 p.m., and 3 p.m.), totaling 750 mL. The irrigation schedule was identical across both experiments. A standardized hydroponic nutrient solution containing essential macro- and micronutrients was applied. The solution included nitrogen, phosphorus, potassium, calcium, magnesium, sulfur, and other elements such as iron, manganese, zinc, copper, boron, and molybdenum. Nutrients were supplied via the irrigation system to ensure uniform distribution. The nutrient solution concentration was adjusted to maintain an electrical conductivity (EC) of 2.0–2.2 dS/m and a pH of 5.8–6.2, optimal for tomato growth. Nutrient solution samples were collected weekly to verify consistency in nutrient concentrations. Adjustments were made as necessary to correct any deviations from target levels.

## 2.2. Sampling

### 2.2.1. Data overview

Two consecutive experiments were conducted to evaluate the proposed methodology under different seasonal conditions. This scheduling allowed for assessing the method under varying natural light conditions due to seasonal changes.

During Exp. #1, 33 dwarf tomato plants were cultivated, and 209 individual TLA measurements were obtained. In Exp. #2, 40 dwarf tomato plants were grown to increase the dataset and enhance statistical robustness. An identical number of 209 TLA measurements were collected.

The consistent number of measurements in both experiments ensured comparability between datasets, while the increased number of plants in Exp. #2 helped address the data imbalance observed in the initial experiment, as detailed in Section 4.2.

2.3. Leaf collection and measurements

Each entry in the research dataset included a reference value of the plant's TLA measured using a LiCOR 3100C leaf area scanner (Fig. 1) and a corresponding set of images derived from a video (Fig. 2). Extending the dataset was a primary concern. Thus, an "onion"-like approach was developed, as explained in the following subsection.

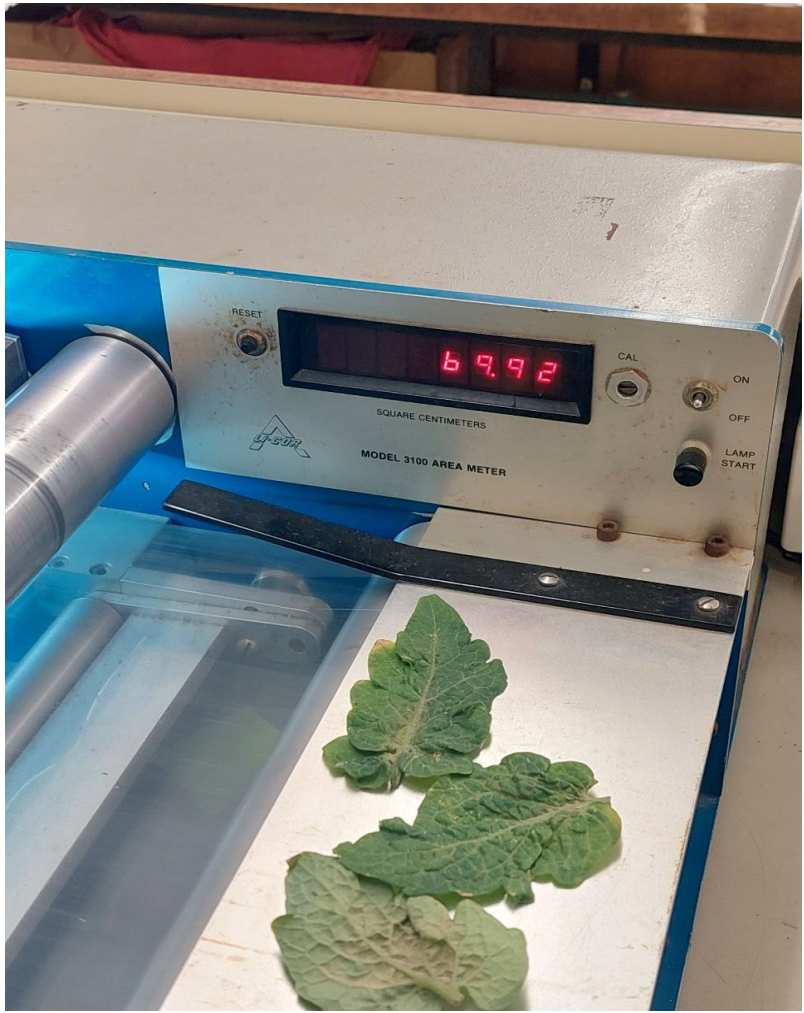

Figure 1. Measuring leaf area with LiCOR 3100C leaf scanner. Leaves are placed on a transparent conveyor belt, where light transmission allows precise surface area calculation, providing accurate ground-truth data for model validation.

2.3.1. Sequential leaf collection (the "Onion" approach)

The "onion" approach is a systematic method for collecting leaf area data by sequentially removing layers of leaves from the plant, akin to peeling an onion (Fig. 3). During the experiment, each plant was assigned a unique identifier and initially photographed in its entirety before any leaf removal, capturing high-resolution videos from multiple angles—including top-down and side perspectives—to create a comprehensive visual record (Fig. 2 and IM.1 in Fig. 4). The outermost leaves, those most distal from the stem, were then carefully removed and immediately placed in labeled bags to prevent desiccation. These removed leaves were measured using the LiCOR 3100C leaf area meter to obtain precise leaf area data (Fig. 1 and LA.1 in Fig. 4). After the first layer of leaves was removed, the plant was re-imaged following the same procedure as before (IM.2 in Fig. 4). This process of leaf removal, measurement, and re-imaging was repeated for each successive layer of leaves, moving inward toward the stem, until all leaves were removed and only the stem remained (IM.1 to IM.5 in Fig. 4).

At each stage, the leaf area was recorded along with the plant's unique identifier and the corresponding layer number in the "onion" process. Finally, a last set of images of the leafless plant structure was captured, and all data were compiled, associating each set of images (IM.n in Fig. 4) with the corresponding TLA measurement (LA.n in Fig. 4).

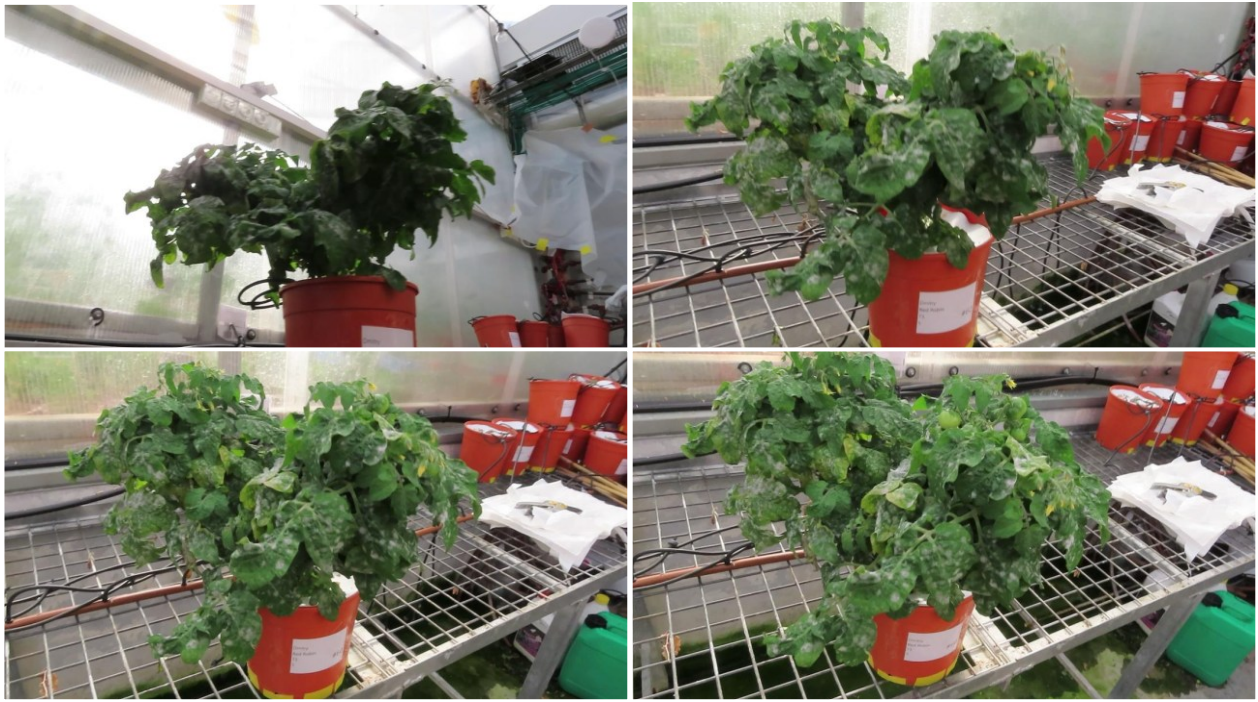

Figure 2. Sample frames extracted from a video of a dwarf tomato plant (link: the video). Image motion was captured from multiple angles and positions to ensure comprehensive 3D reconstruction of complex canopies.

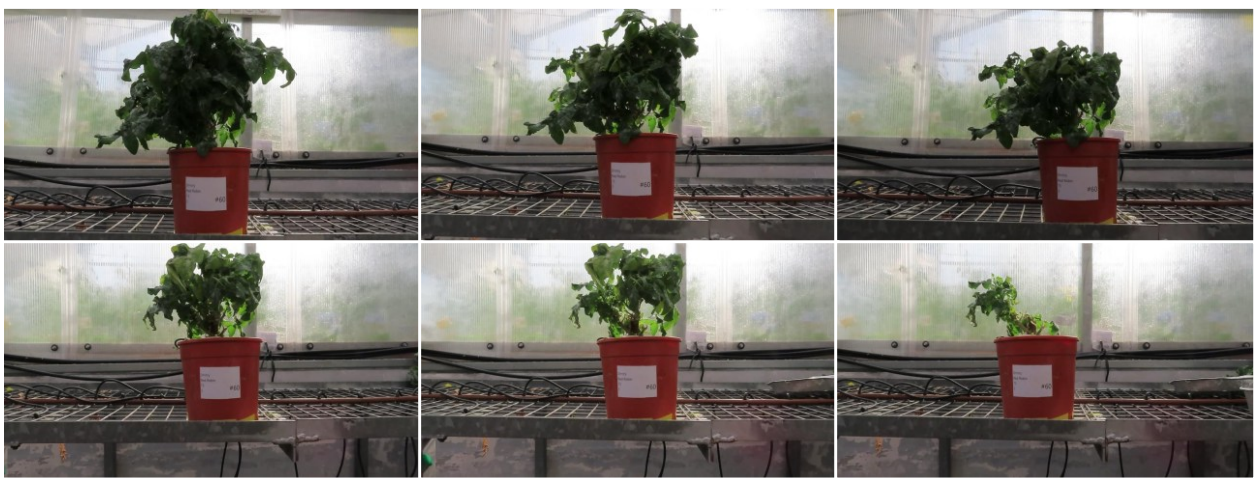

Figure 3. (a – f) Images showing sequential leaf removal stages ("onion" approach) in a dwarf tomato plant from (a) a plant with a full canopy to (c) a plant with a partial removal of 40% of the canopy, and (f) the same plant after removing 90% of its leaves. This method enables the TLA dataset to be increased in this way.

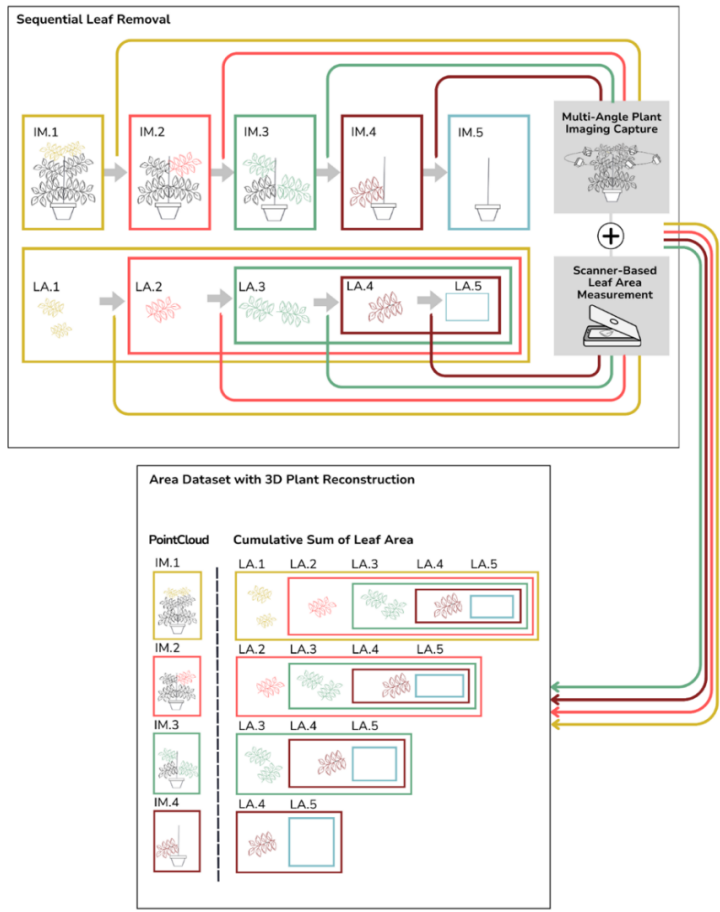

Figure 4. Schematic representation of the "onion" approach, illustrating the sequential removal of leaf layers and associated imaging and measurement steps.

This approach has several advantages. First, it provides multiple data points per plant, improving the training dataset for machine learning models. It also captures a range of canopy densities and structures, enhancing the robustness of the 3D reconstruction algorithms. It is also important that the imaging process is non-destructive; it does not harm the plant, allowing for an accurate representation of the plant at each stage.

## 2.4. Imaging techniques

### 2.4.1. Video capture protocol

To capture high-resolution videos suitable for 3D reconstruction, a Canon PowerShot SX70 HS digital camera was used. The camera's sensor is a CMOS sensor with a 20.3-megapixel resolution. The lens focal length is 4.3 mm (equivalent to 21 mm in 35 mm format). Manual focus was used as it is important for the photogrammetry process. The resolution of the videos was 1920x1080.

To create a comprehensive dataset for 3D reconstruction, each plant was imaged from multiple angles using a systematic procedure. The videos were recorded in the greenhouse environment. Additional light was occasionally turned on to achieve consistent image quality throughout all the videos, and it did not depend on the light conditions in the greenhouse.

We recorded the plant from all possible angles by slowly moving the camera, shooting the video around the plant in horizontal and vertical directions, and rotating the camera to achieve maximum coverage. The videos were typically 40-60 seconds long, depending on the plant size and the shooting process condition (Fig. 3).

To maintain consistency in the dataset, the same imaging protocol was applied to all plants and at each stage of the "onion" approach.

*2.4.2. Frame extraction for photogrammetry*

After recording, 500 evenly distributed individual frames were extracted from each video using a custom Python script to serve as input for the 3D reconstruction process.

*2.5. Image processing and point cloud generation*

*2.5.1. 3D reconstruction with Meshroom*

The 500 extracted frames per plant were imported into Meshroom, retaining metadata (EXIF data) to assist with camera calibration. During camera initialization, the software automatically detected the camera model and extracted intrinsic parameters. For feature extraction, the SIFT (Scale-Invariant Feature Transform) algorithm was used to achieve robust feature detection under varying scales and orientations (J. Wu et al., 2013). Default settings were applied, adjusting the number of features detected if required. A brute-force matcher with cross-checking was used for feature matching to find correspondences between features in different images. Epipolar constraints were applied to filter out incorrect matches. The software then incrementally reconstructed the scene using Structure-from-Motion (SfM) by estimating camera poses and 3D points. Global bundle adjustment was performed to optimize camera parameters and 3D point positions, minimizing reprojection errors, resulting in a sparse raw point cloud representing the 3D structure of the plant along with estimated camera positions. For this study, the pipeline was terminated after the SfM step to focus on the raw point cloud without generating dense depth maps or textured meshes at this stage. Finally, the raw point cloud and camera position data were exported as .ply (Polygon File Format) files for compatibility with subsequent processing tools.

*2.5.2. Point cloud refinement steps*

After obtaining the raw point cloud, several refinement steps were necessary to prepare the data for mesh reconstruction and subsequent analysis.

Figure 5 shows a schematic representation of the point cloud refinement pipeline.

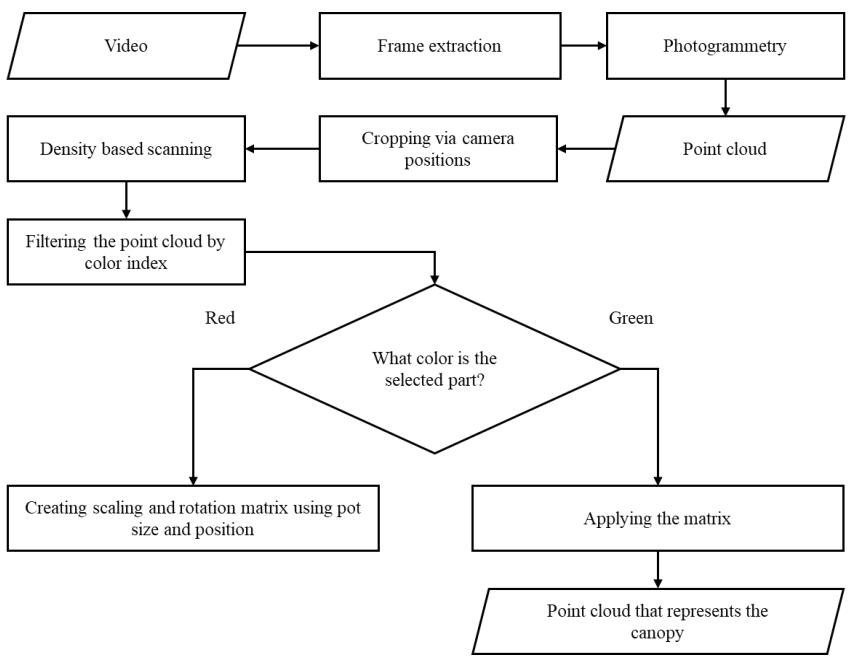

Figure 5. Pipeline for generating and refining 3D point clouds from RGB images.

2.5.2.1. Cropping the point cloud

The point cloud was cropped to retain only the points within a defined cubical region centered around the plant. The cube's dimensions were determined based on the camera positions extracted from the SfM process. A script was developed to automate the cropping process.

2.5.2.2. Noise reduction with DBScan clustering

DBScan (Density-Based Spatial Clustering of Applications with Noise) was applied to identify and remove noise. Values for epsilon, representing the maximum distance between points considered part of the same cluster, and minimum samples, the minimum number of points required to form a dense region, were selected empirically by testing different values and observing their effects on retaining meaningful plant structures while removing sparse noise points. The chosen values effectively removed isolated points and small clusters unlikely to represent actual plant features.

2.5.2.3. Color filtering to isolate plant material

RGB color values were converted to green index ($I_g$) to differentiate between plant material and non-plant elements:

$$I_g = \frac{G - R}{R + G} \qquad (1)$$

where G and R are the green and red band digital number values, respectively. An $I_g$ value higher than –0.02 was considered plant material and passed as parts of the "canopy." Those points with $I_g < -0.25$ were regarded as non-plant material and passed as parts of the "pot." The points with $I_g$ between –0.25 and –0.02 were dismissed. The above $I_g$ thresholds were set empirically after manually inspecting the point cloud images.

2.5.2.4. Scaling and rotation based on pot dimensions

Due to its known dimensions and orientation, the pot was used as a reference object. Points classified as "pot" from the color filtering step were isolated. Using the Least Squares Method, a plane was fitted to the pot points, ensuring that the pot's top surface corresponded to the XY plane (Z=0). Then, a circle was fitted to the pot points on the fitted plane to determine its center and radius. The known pot diameter (15 cm) was used to scale the point cloud accurately. A scaling matrix was applied to adjust the point cloud dimensions to match real-world measurements. Scaling factors were calculated based on the ratio of the known pot size to the measured size in the point cloud. The point cloud was also rotated to align the pot base with the XY-plane and center the plant along the Z-axis.

Transformations were implemented using a custom Python script for efficient computation. The transformed point clouds were exported and visualized to confirm correct scaling and alignment.

### 2.6. Mesh reconstruction algorithms

Four reconstruction algorithms were employed to convert the refined point clouds into usable 3D meshes: Alpha Shape, Marching Cubes, Poisson's 3D Reconstruction, and Ball Pivoting. Each algorithm offers unique advantages and challenges, and exploring multiple methods ensures the robustness of the mesh generation process. Also, each algorithm has a specific parameter that was exhausted to find the best reconstruction model.

Table 2 provides the parameters of the four reconstruction algorithms.

### 2.6.1. Alpha shape

The Alpha Shape algorithm is a generalization of the convex hull concept that allows for reconstructing concave shapes from a set of points. The algorithm constructs a shape that tightly wraps around the point cloud based on a parameter α (alpha), which determines the level of detail and concavity of the resulting mesh. A comprehensive set of 29 alpha values was explored to determine the optimal parameter for accurately reconstructing the plant meshes.

Table 2. Exhausted parameters for each shape reconstruction algorithm.

| Algorithm | Parameter | Exhausted options |
|---|---|---|
| *Alpha shape* | Alpha value | 0.001, 0.005, 0.01, 0.05, 0.075, 0.1, 0.2, 0.3, 0.4, 0.5, 0.6, 0.7, 0.8, 0.9, 1, 1.25, 1.5, 1.75, 2, 3, 4, 5, 7.5, 10, 15, 20, 25, 50, 100, 1000 |
| *Marching cubes* | Threshold value | 0.1, 0.2, 0.3, 0.4, 0.5, 0.6, 0.7, 0.8, 0.9, 1, 1.1, 1.2, 1.3, 1.4, 1.5, 1.6, 1.7, 1.8, 1.9, 2, 2.1, 2.2, 2.3, 2.4, 2.5, 2.6, 2.7, 2.8, 2.9, 3, 4, 5, 6, 7, 8, 9, 10, 11, 12, 13, 14, 15 |
| *Poisson's* | Depth | 8, 9, 10, 11, 12, 13, 14, 15 |
| *Ball pivoting* | Ball radii | [0.05, 0.1, 0.2, 0.4], [0.1, 0.2, 0.4, 0.8], [0.152, 0.3, 0.61, 1.2], [0.2, 0.4, 0.8, 1.6], [0.25, 0.5, 1.0, 2.0], [0.3, 0.6, 1.2, 2.4], [0.35, 0.7, 1.4, 2.8], [0.4, 0.8, 1.6, 3.2], [0.45, 0.9, 1.8, 3.6], [0.5, 1.0, 2.0, 4.0] |

### 2.6.2. Marching cubes

The Marching Cubes algorithm is a high-resolution isosurface extraction method. It operates by dividing the 3D space into a grid of cubes (voxels) and constructing surfaces (meshes) that represent points of equal value (isosurfaces) within the data. A threshold value defines the isosurface level to extract. Adjusting the threshold alters the detail and extent of the reconstructed surface. A total of 41 threshold values ranging between 0.1 and 15 were tested to find the most suitable for plant mesh reconstruction. The reconstruction procedure was implemented using the PyMCubes library for Python (Pablo Márquez Neila, 2023), which efficiently implements the Marching Cubes algorithm.

### 2.6.3. Poisson's 3D reconstruction

Poisson's 3D Reconstruction is a method that formulates surface reconstruction as a spatial Poisson problem, integrating point cloud data to produce smooth, watertight surfaces. The depth parameter controls the octree depth used in the reconstruction, affecting the resolution of the mesh. Higher depth values result in finer details but increase computational load. Integer depth values between 8 and 15 were tested.

The algorithm was employed using the Open3D library's Poisson reconstruction function.

### 2.6.4. Ball pivoting

The Ball Pivoting Algorithm (BPA) reconstructs surfaces by simulating the rolling of a virtual ball over the point cloud, connecting points where the ball touches to form triangles. When using ball pivoting, multiple radii can be specified to capture features of different scales within the point cloud. 10 sets of 4 radii values were exhausted.

The algorithm was employed using the Open3D library.

### 2.7. Feature extraction from meshes

After generating the meshes, various geometric parameters were calculated to serve as features for machine learning models.

Table 3 presents the list of all calculated mesh parameters.

Table 3. The list of parameters calculated from the 3D meshes. Formulas for the simple geometrical parameters are provided. In other cases, parameters were computed based on the mesh shapes.

| Parameter | Description | Formula | |
|---|---|---|---|
| Height (H), cm | The vertical d extent of the mesh along the Z-axis | $H = z_{max} - z_{min}$ | (2) |
| Length (L), cm | The largest dimension of the mesh projected on XY-plane | $L = x_{max} - x_{min}$ | (3) |
| Width (W), cm | The smallest dimension of the mesh projected on XY-plane | $W = y_{max} - y_{min}$ | (4) |
| Aspect ratio (AR), cm·cm−1 | The ratio of length to width | $AR = L / W$ | (5) |
| Volume (V), cm³ | The total enclosed volume of the mesh | | |
| Surface area (A), cm² | The total area of all the mesh's faces | | |
| Bounding box area ($A_b$), cm² | The area of the minimal bounding box that encloses the mesh when projected on horizontal XY-plane | | |
| Bounding box volume ($V_b$), cm³ | The volume of the minimal bounding box that encloses the mesh | | |
| Number of individual mesh components (NC) | The count of disconnected components within the mesh | | |

### 2.8. Machine learning models and statistical analysis

#### 2.8.1. t-SNE analysis

To account for the fact that different cultivars were used in experiments carried out in different seasons and that the data was collected with a specific sequential approach, we used a t-distributed Stochastic Neighbor Embedding (t-SNE) analysis. The t-SNE algorithm is a non-linear dimensionality reduction technique well-suited for visualizing high-dimensional data in a lower-dimensional space (typically 2D or 3D) (Van Der Maaten & Hinton, 2008).

The dataset comprised all mesh parameters extracted from the 3D reconstructions, including height, length, width, volume, surface area, aspect ratio, bounding box dimensions, and the number of mesh components. Other parameters included in the analysis were cultivar, treatment during the experiment (pruning intensity), and sequence number according to the "onion" approach.

#### 2.8.2. Regression models

Several regression models were evaluated to predict leaf area from mesh parameters.

*Multivariable linear regression (MLR)* assumes a linear relationship between independent variables (mesh parameters) and the dependent variable (leaf area).

*Lasso regression (Lasso)* adds an L1 regularization term to the loss function, promoting sparsity in the coefficients and performing feature selection (Fonti et al., 2017).

*Ridge regression (Ridge-Reg)* incorporates an L2 regularization term to penalize large coefficients, reducing multicollinearity effects.

*Elastic net regression (ENR)* combines L1 and L2 regularization, balancing between *Lasso* and *Ridge-Reg* benefits.

*Random forest (RF)* is an ensemble of decision trees that improves predictive accuracy and controls overfitting.

*Extreme gradient boosting* (*XGBoost*) is an optimized gradient boosting algorithm that builds additive models forward stage-wise (Chen & Guestrin, 2016).

*Multilayer Perceptron (MLP)* is a feedforward artificial neural network capable of modeling complex nonlinear relationships.

### 2.8.3. Data pre-processing

Features were standardized by removing the mean and scaling to unit variance. This ensures that all features contribute equally to the model training.

Fig. 6 illustrates the steps completed for the data preparation.

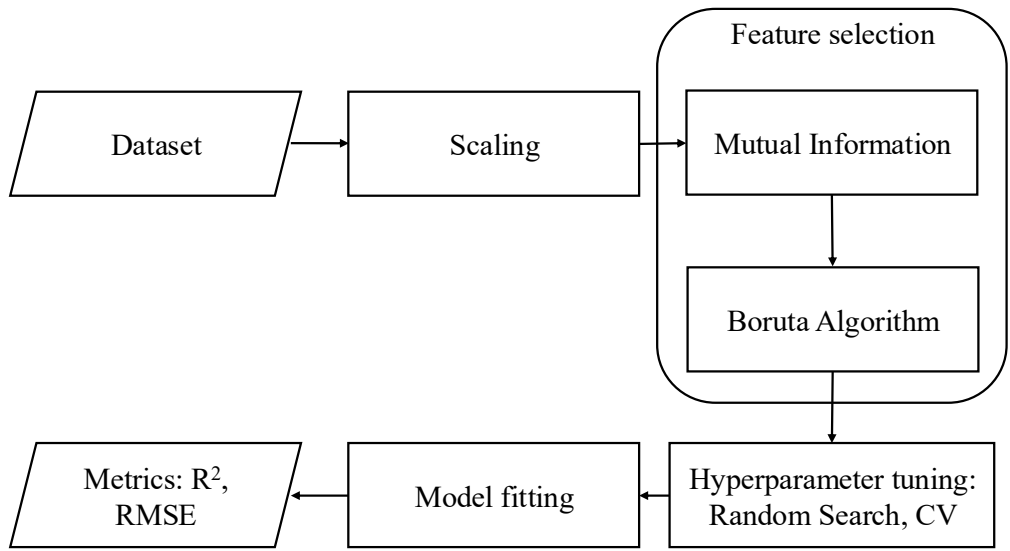

Figure 6. Data preparation pipeline. This pipeline includes standardization of mesh parameters, feature selection through mutual information and Boruta algorithms, and model training with various regression techniques.

### 2.8.4. Feature selection methods

Feature selection was executed by estimating mutual information (MI). It measures the dependency between features and the target variable. Features with MI scores exceeding 0.1 were retained. Additionally, the Boruta algorithm was used to refine feature selection (Kursa et al., 2010). It was employed via the *BorutaPy* package in Python.

### 2.8.5. Model training and validation

For all ML algorithms, cross-validation-based random search was used to fine-tune hyperparameters. Data was split in two distinct ways: 1) the whole dataset, including results from two experiments, was stratified and then split into train and test subsets as 70% to 30%. The stratification implies that, in this case, train and test subsets always have close distributions; 2) the dataset was split into halves by the experiment number. This made it possible to use whole experiment datasets as unseen data. In any case, train sets were split into six folds to perform K-fold cross-validation. The best model of these six folds was applied to the test subset, and then the metrics of the prediction model were measured. $R^2$ and the mean average error (MAE) were used to assess the quality of the models (Fig. 7).

*2.8.6. Software and Tools*

The analysis was done using Python 3.10. Scikit-learn library was used for machine learning models and feature selection. XGBoost library was used for gradient boosting models. Analysis was conducted on HUJI's cluster computation center.

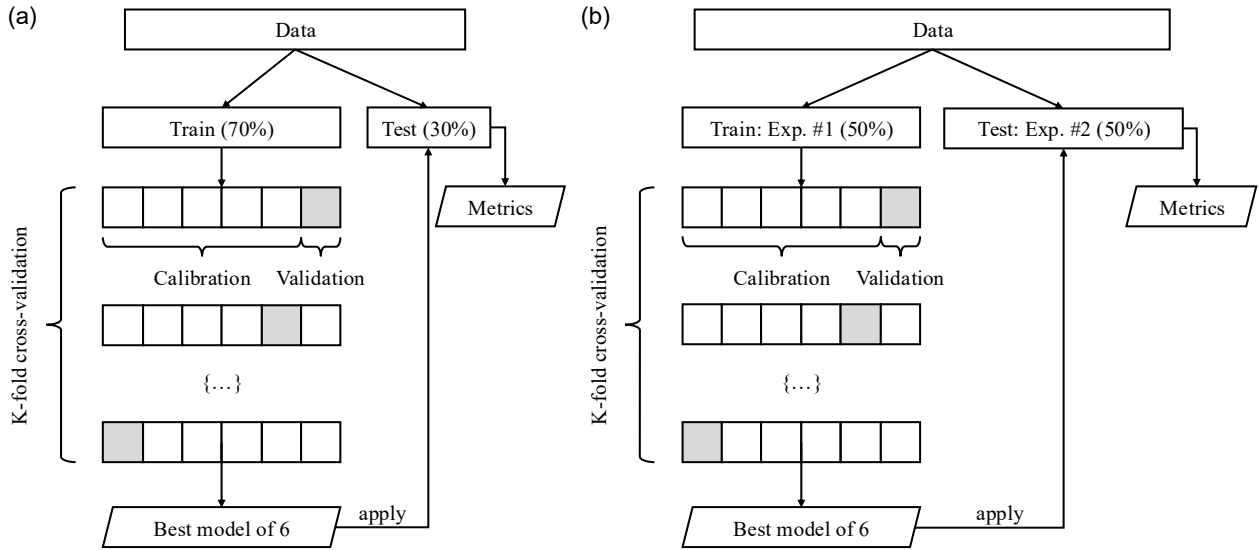

Figure 7. Data split and cross-validation scheme for ML regression analysis. (a) A 70/30 split of the whole dataset from the two experiments and (b) data obtained from Exp. #1 to reproduce Exp. #2.

## 3. Results

The two experiments provided significantly different datasets of TLA values (Table 4 and Fig. 8). In Exp. #1, the distribution was skewed towards smaller leaf area values, with a higher frequency of measurements between 500 $cm^2$ and 1,500 $cm^2$. Fewer samples represent larger leaf areas exceeding 3,000 $cm^2$.

Table 4. Characteristic data distribution of total leaf area (TLA) values for Exp #1 and Exp. #2. Min, Max, Mean, Median, and Std are the population sample's minimum, maximum, mean, median, and standard deviation leaf areas (in $cm^2$), respectively.

| TLA | Min ($cm^2$) | Max ($cm^2$) | Mean ($cm^2$) | Median ($cm^2$) | Std ($cm^2$) |
|---|---|---|---|---|---|
| Exp. #1 | 150.3 | 5250.7 | 1245.6 | 1050.4 | 875.2 |
| Exp. #2 | 200.5 | 6300.9 | 1900.8 | 1600.2 | 1050.7 |
| Combined | 150.3 | 6300.9 | 1573.2 | 1316.8 | 983.5 |

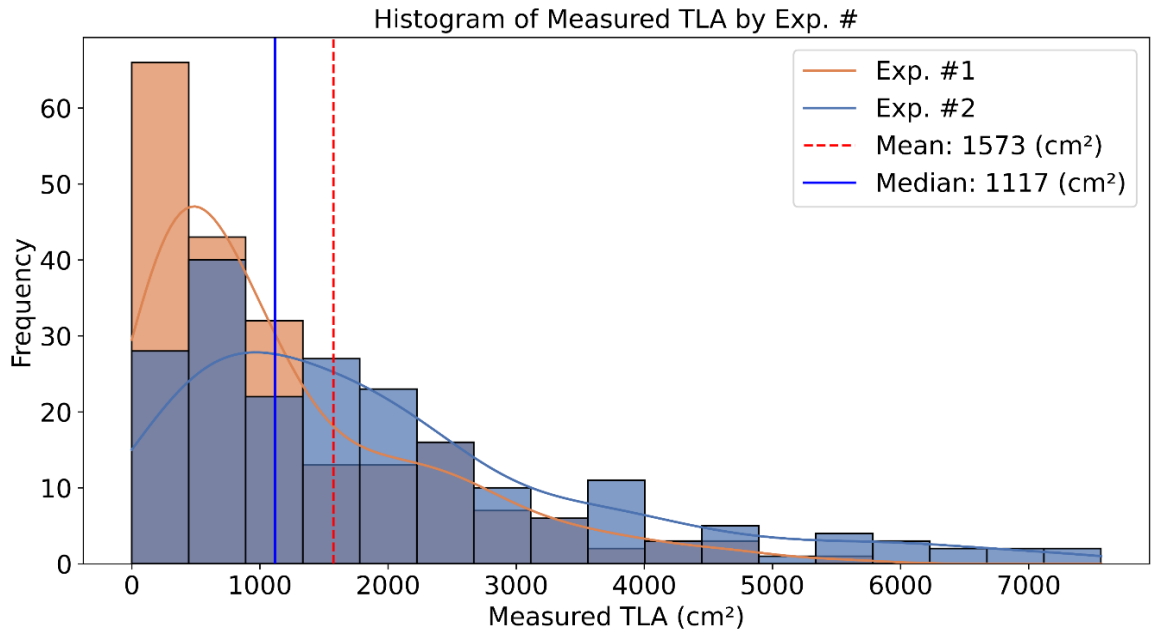

Figure 8. Histogram of total leaf area (TLA) measurements across two experiments. Exp. #1 (blue) shows a skew towards smaller TLAs, while Exp. #2 (orange) including much larger TLAs. The red dashed line indicates the combined mean TLA, and the solid blue line represents the median TLA.

In Exp. #2, the distribution was more uniform, with increased measurements in the higher leaf area range (2,000 $cm^2$ to 5,000 $cm^2$). This reflects a deliberate effort to capture more data from plants with abundant foliage.

The combined dataset shows a broader distribution of TLAs, enhancing the dataset's representativeness across different plant sizes and canopy densities.

### 3.1. Data Clusters

The t-SNE shows that data points are distributed without forming distinct clusters or groups (Fig. 9). This dispersion suggests no inherent subgroups within the dataset based on the input features. This stays true for all input parameters, including cultivar, treatment, and experiment number. It indicates that these experimental variables did not significantly influence the mesh parameters or the measured TLA. The overlapping distributions of different cultivars and treatments (not shown) suggest that the plant's morphological features captured by the mesh parameters are consistent across these variables. The absence of clusters related to experimental variables supports the pooling of data from different cultivars, experiments, and treatments for model training. It validates the assumption that the relationships between mesh parameters and leaf area are consistent across the dataset.

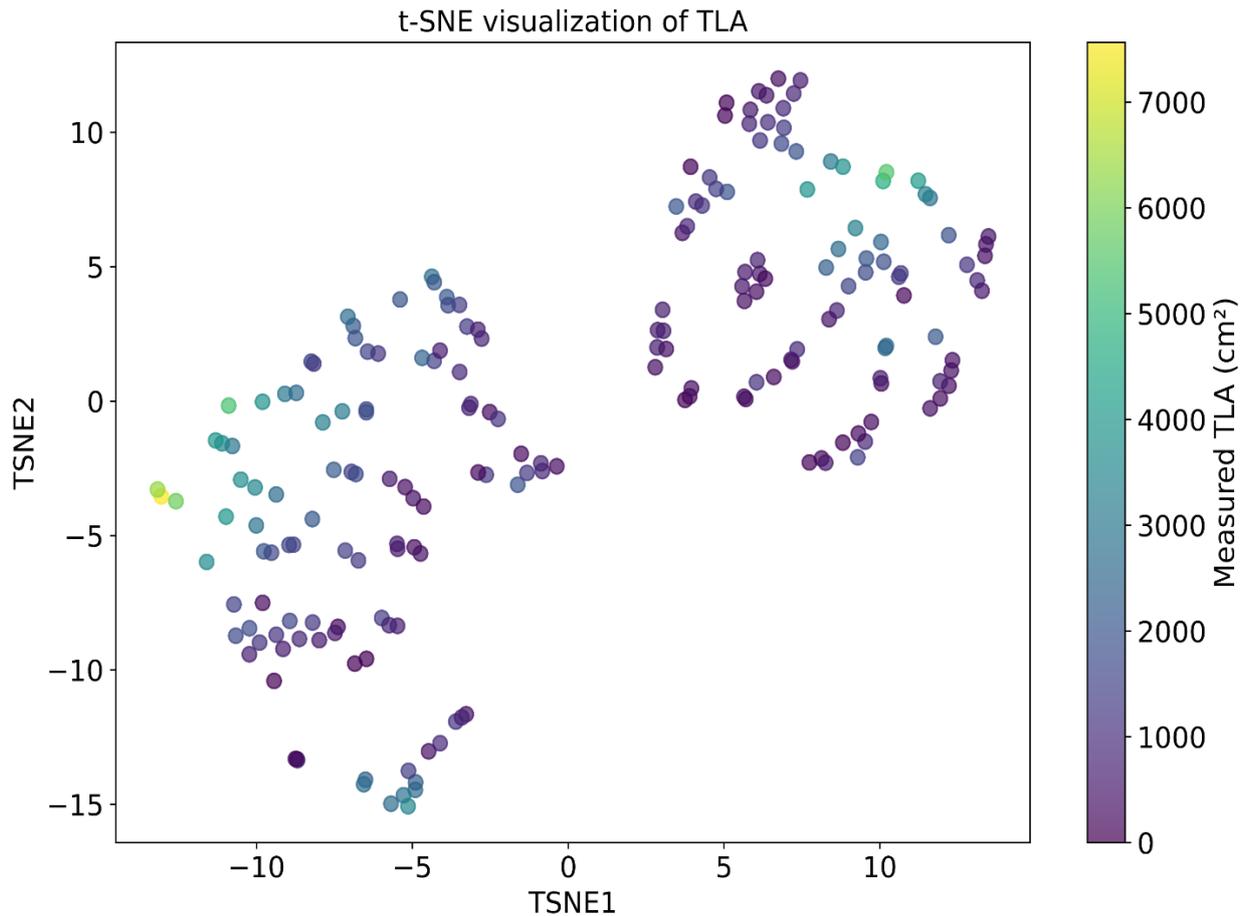

Figure 9. The t-SNE plot of the dataset is colored by the measured total leaf area (TLA in cm2). Each point represents a measurement (a plant at a specific stage in the "onion" approach sequence). The points are plotted in a two-dimensional space resulting from the t-SNE algorithm. The color gradient corresponds to the measured TLA, with a continuous scale from purple (smaller TLAs) to yellow (larger TLAs).

### 3.2. Regression model evaluation

#### 3.2.1. Model performance metrics

Table 5 presents each model's best $R^2$ and MAE (in cm$^2$) values. Using the whole dataset split 70/30 into train/test sets, *XGBoost* achieved the highest $R^2$ (0.8) and lowest MAE (489 cm$^2$; 31%) when applied to alpha-shapes created with *α*=3, indicating superior predictive accuracy above other shape parameters and ML algorithms. *Random Forest (RF)* also performed well but was slightly less accurate than *XGBoost*.

Linear models (*MLR*, *Lasso*, *Ridge-Reg*, and *Elastic Net*) showed consistent performance, suggesting linear relationships between features and TLA. *MLP* generally had lower performance, possibly due to overfitting or insufficient data volume for deep learning methods.

When training and testing across different experiments, model performance decreased significantly, yielding the following best results:

- Train on Exp. #1 and test on Exp. #2 yielded an $R^2$ = 0.55 and MAE = 739 cm$^2$ (*Lasso*)
- Train on Exp. #2 and test on Exp. #1 yielded an $R^2$ = 0.56 and MAE = 579 cm$^2$ (*MLP*)

Table 5. Performance metrics of the ML models. Only the best scores per model are presented.

| Prediction | | Whole dataset (70/30) | | Train Exp #1 / Test Exp #2 | | Train Exp #2 / Test Exp #1 | |
|---|---|---|---|---|---|---|---|
| Reconstruction algorithm | Model | $R^2$ | MAE ($cm^2$) | $R^2$ | MAE ($cm^2$) | $R^2$ | MAE ($cm^2$) |
| Alpha | MLR | 0.71 | 587 | 0.55 | 745 | 0.54 | 532 |
| | Lasso | 0.72 | 515 | 0.55 | 739 | 0.54 | 590 |
| | Ridge-Reg | 0.72 | 535 | 0.53 | 759 | 0.53 | 609 |
| | ENR | 0.72 | 561 | 0.53 | 746 | 0.54 | 578 |
| | RF | 0.72 | 496 | 0.50 | 665 | 0.43 | 638 |
| | XGBoost | 0.80 | 489 | 0.52 | 773 | 0.48 | 602 |
| | MLP | 0.69 | 510 | 0.54 | 737 | 0.56 | 579 |
| Ball pivoting | MLR | 0.68 | 506 | 0.54 | 739 | 0.48 | 615 |
| | Lasso | 0.67 | 503 | 0.54 | 732 | 0.52 | 599 |
| | Ridge-Reg | 0.69 | 521 | 0.53 | 747 | 0.48 | 627 |
| | ENR | 0.71 | 562 | 0.55 | 726 | 0.52 | 578 |
| | RF | 0.68 | 496 | 0.53 | 768 | 0.38 | 663 |
| | XGBoost | 0.66 | 607 | 0.51 | 651 | 0.44 | 649 |
| | MLP | 0.71 | 477 | 0.52 | 631 | 0.47 | 654 |
| Marching cubes | MLR | 0.71 | 456 | 0.44 | 694 | 0.39 | 688 |
| | Lasso | 0.70 | 495 | 0.46 | 689 | 0.43 | 641 |
| | Ridge-Reg | 0.69 | 568 | 0.46 | 786 | 0.45 | 660 |
| | ENR | 0.71 | 573 | 0.44 | 686 | 0.42 | 680 |
| | RF | 0.69 | 600 | 0.48 | 662 | 0.32 | 702 |
| | XGBoost | 0.67 | 587 | 0.50 | 665 | 0.47 | 645 |
| | MLP | 0.71 | 459 | 0.51 | 648 | 0.54 | 583 |
| Poisson | MLR | 0.61 | 645 | 0.50 | 786 | 0.41 | 677 |
| | Lasso | 0.62 | 605 | 0.51 | 769 | 0.41 | 679 |
| | Ridge-Reg | 0.66 | 638 | 0.49 | 787 | 0.42 | 670 |
| | ENR | 0.65 | 532 | 0.50 | 780 | 0.39 | 600 |
| | RF | 0.65 | 638 | 0.49 | 811 | 0.25 | 743 |
| | XGBoost | 0.62 | 650 | 0.50 | 671 | 0.40 | 653 |
| | MLP | 0.66 | 491 | 0.50 | 773 | 0.41 | 633 |

### 3.2.2. Feature importance analysis

Using the *XGBoost* model, we analyzed the importance of features based on the gain metric. Mutual information and Boruta were applied to the analysis, leaving out all features except height, width, and surface area. The most predictive feature, indicating vertical growth (height), is a strong indicator of TLA (Fig. 10). The surface area reflects the complexity and extent of the plant canopy, while canopy width contributes to understanding the plant's dimensions.

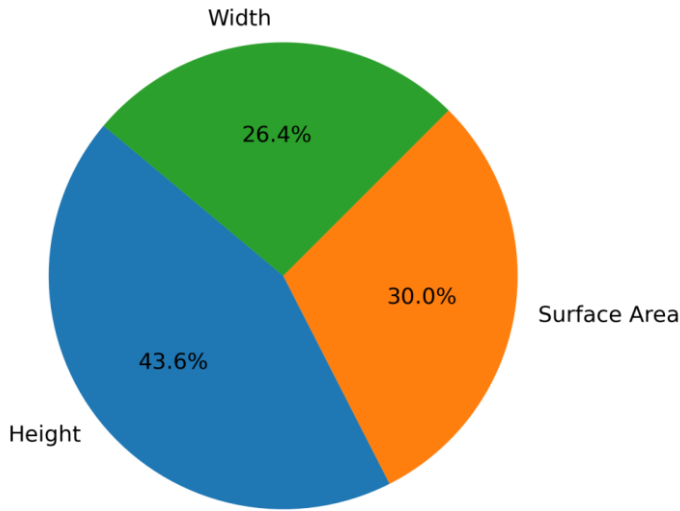

Figure 10. Feature importance analysis from the best-performing XGBoost model. Height, width, and surface area of 3D-reconstructed meshes emerged as the most predictive features for estimating TLA after applying mutual information and Boruta selection methods.

To assess model accuracy, we plotted predicted against actual TLAs (Fig. 11a). Data points closely align along the 1:1 line, with an average bias of 60 cm$^2$ (only 3.8%). Slight deviations at higher leaf areas suggest reduced accuracy for relatively large plants, likely due to fewer samples in this range. Residuals are distributed around zero, with no apparent patterns, indicating no systematic bias (Fig. 11b). Variability increases at higher leaf areas, consistent with observed deviations.

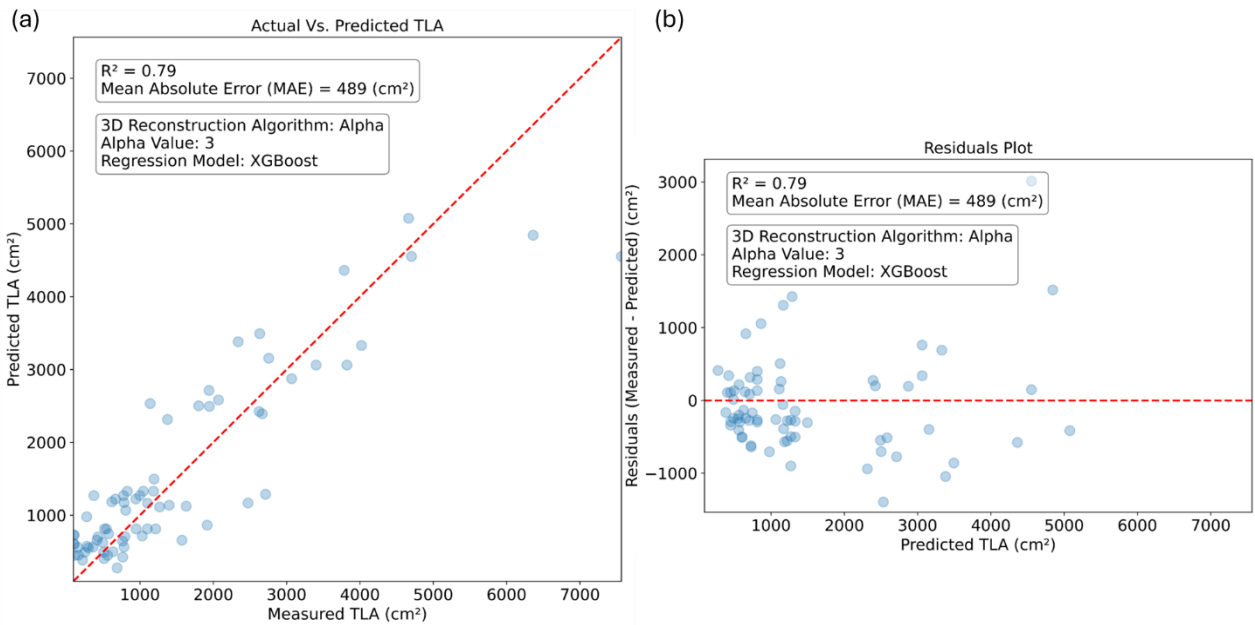

Figure 11. Model performance evaluation of XGBoost predictions. (a) Scatter plot comparing predicted and actual TLA values along the 1:1 line. (b) Residual plot showing minimal bias and random error distribution, with slight variability at higher TLA values.

# 4. Discussion

## 4.1. Addressing data imbalances

The initial data analysis from Exp. #1 revealed a significant imbalance in the distribution of total leaf area measurements, i.e., the "onion" approach inherently generated more measurements with smaller leaf areas as leaves were progressively removed, resulting in numerous samples representing the diminishing foliage. This led to an overrepresentation of smaller leaf area values and an underrepresentation of larger ones.

To address this imbalance in Exp. #2, strategic measures were implemented. Sampling was deliberately focused on larger canopies to increase the number of measurements in the higher leaf area range. The number of iterations in the leaf removal process was modified. Instead of peeling layers down to minimal residuals, the process was halted earlier for some plants to retain larger leaf areas in the dataset.

The inclusion of more samples with larger leaf areas from the second experiment made the dataset more representative of actual plant growth scenarios, especially for mature plants. However, it is essential to note that, despite the efforts, a significant imbalance remained due to natural biological limitations (the growth rate of plants imposes natural constraints on achieving large total leaf areas within the experimental timeframe) and methodological constraints (the "onion" approach's design inherently produced more data points with smaller leaf areas as leaves are removed).

## 4.2. Interpretation of model performance

Combining *XGBoost* with the Alpha shape reconstruction algorithm ($\alpha = 3$) yielded the highest predictive accuracy in estimating total leaf area from 3D mesh-derived features. Several factors contributed to this superior performance.

First, *XGBoost*, a gradient-boosting framework, excels at modeling complex nonlinear interactions between features and the target variable. Although linear models show strong results, the nature of plant morphology and its representation through mesh parameters involves intricate relationships that are not purely linear. *XGBoost* ability to handle these complexities allows it to extract more informative patterns from the data. *XGBoost* is also less sensitive to outliers compared to models like linear regression. In datasets with measurement errors or anomalies, *XGBoost* can maintain performance by assigning less weight to problematic data points during training. Finally, *XGBoost* incorporates regularization parameters that prevent overfitting.

Alpha Shape with $\alpha = 3$ effectively captures the fine detail of the plant structure, including individual leaves and intricate canopy shapes. The meshes produced are neither overly simplified nor excessively complex. This balance provides meaningful features without introducing unnecessary noise that could confuse the regression model.

In contrast, Poisson's 3D reconstruction, even with an optimal depth setting of 11, resulted in lower model performance. Poisson's reconstruction tends to generate smooth, watertight surfaces, which, while useful in some applications, can obscure critical plant details. The resulting over-smoothing merges fine structures, diminishing the distinctiveness of mesh parameters, specifically surface area. In addition, the algorithm is more sensitive to variations in point cloud density and noise. Any imperfections in the point cloud can significantly impact the quality of the reconstructed mesh.

However, although selecting a reconstruction algorithm is essential, it is usually only responsible for 0.1–0.15 of the variability in total leaf area. Feature importance analysis states that plant dimensions are the main drivers of the model's predictive power. It implies that the reconstruction algorithm's ability to correctly assume the plant's dimensions is most important, and all tested algorithms succeed. Further improvement hinges on the parameters accurately reflecting the plant's physical characteristics.

*4.3. Limitations of the study*

Despite the efforts to balance the dataset, there remained a higher frequency of measurements with smaller leaf areas due to the nature of the "onion" approach. The model's predictive accuracy decreases for the larger values, as evidenced by increased residuals in that range. Future studies could address this by employing sophisticated sampling strategies or data augmentation techniques, such as SMOTE (Synthetic Minority Over-sampling Technique) (Chawla et al., 2002) or adaptive synthetic sampling, to generate synthetic data for larger leaf areas and balance the dataset.

Although the process is highly automated, one step in the data processing pipeline requires manual intervention – i.e., scaling. For each plant, the researcher needs to search for the top circle of the pot and extract it so that a scaling matrix is calculated. This is fixable by introducing a new type of scaling and rotation reference that is repeatedly extractable automatically, but for now, it could limit the scalability and automation potential.

Despite these limitations, the research presents a TLA prediction model created upon a highly automated image collection and processing procedure that yields highly accurate results compared to traditional methods. Unlike previous studies, our method enables assessing TLAs of relatively highly bushy plants without the use of a motion platform. Our solution may be well-suited for use in autonomous systems.

## 5. Future work and scalability

The described method can be tested in urban farming environments, including vertical farms. The model's setup does not seem to prevent it from performing well under varying conditions. However, this implementation's effectiveness should be further assessed.

If commercial large-scale use is considered, the interfaces or APIs should be developed to integrate our method into existing agricultural software platforms seamlessly.

Integrating 3D reconstruction techniques with advanced machine learning models significantly advances non-destructive leaf area estimation. The different performance of reconstruction algorithms and regression models underscores the importance of selecting appropriate computational methods tailored to the specific characteristics of plant morphology. While limitations exist, the study provides a robust foundation for future research and practical applications in automated pruning and urban farming efficiency. By addressing the challenges and pursuing the opportunities outlined, this approach has the potential to transform agricultural practices and contribute to sustainable food production in urban environments.

**Acknowledgments**

The authors thank Yedidya Harris for helping maintain the plants in the greenhouse and Gabriel Mulero for helping with data collection and measurements. D.U. acknowledges personal funding from the SCE-HUJI Scholarship Program for Ph.D. students and the Scholarships for Research Students from the Ministry of Aliyah and Integration of Israel. D.U. is a Ph.D. student at the Faculty of Agriculture, Food, and Environment under the supervision of D.H. and C.G.

**Table captions**

Table 1. Growing conditions in the two experiments (Exp. #1 and Exp. #2).

Table 2. Exhausted parameters for each shape reconstruction algorithm.

Table 3. The list of parameters calculated from the 3D meshes. Formulas used to calculate the parameters are provided, if possible. In other cases, the parameters were computed based on the mesh shapes.

Table 4. Characteristic data distribution of total leaf area (TLA) values for Exp #1 and Exp. #2. Min, Max, Mean, Median, and Std are the population sample's minimum, maximum, mean, median, and standard deviation leaf areas (in $cm^2$), respectively.

Table 5. Performance metrics of the regression models. Only the best scores obtained, exhausting all parameter options, are presented for each regression model.

**Figure captions**

Figure 1. Measuring leaf area with LiCOR 3100C leaf scanner. Leaves are placed on a transparent conveyor belt, where light transmission allows precise surface area calculation, providing accurate ground-truth data for model validation.

Figure 2. Sample frames extracted from a video of a dwarf tomato plant (link: the video). Image motion was captured from multiple angles and positions to ensure comprehensive 3D reconstruction of complex canopies.

Figure 3. (a – f) Images showing sequential leaf removal stages ("onion" approach) in a dwarf tomato plant from (a) a plant with a full canopy to (c) a plant with a partial removal of 40% of the canopy, and (f) the same plant after removing 90% of its leaves. This method enables the TLA dataset to be increased in this way.

Figure 4. Schematic representation of the "onion" approach, illustrating the sequential removal of leaf layers and associated imaging and measurement steps.

Figure 5. Pipeline for generating and refining 3D point clouds from RGB images.

Figure 6. Data preparation pipeline. This pipeline includes standardization of mesh parameters, feature selection through mutual information and Boruta algorithms, and model training with various regression techniques.

Figure 7. Data split and cross-validation scheme for ML regression analysis. (a) A 70/30 split of the whole dataset from the two experiments and (b) data obtained from Exp. #1 to reproduce Exp. #2.

Figure 8. Histogram of total leaf area (TLA) measurements across two experiments. Exp. #1 (blue) shows a skew towards smaller TLAs, while Exp. #2 (orange) including much larger TLAs. The red dashed line indicates the combined mean TLA, and the solid blue line represents the median TLA.

Figure 9. The t-SNE plot of the dataset is colored by the measured total leaf area (TLA in $cm^2$). Each point represents a measurement (a plant at a specific stage in the "onion" approach sequence). The points are plotted in a two-dimensional space resulting from the t-SNE algorithm. The color gradient corresponds to the measured TLA, with a continuous scale from purple (smaller TLAs) to yellow (larger TLAs).

Figure 10. Feature importance analysis from the best-performing XGBoost model. Height, width, and surface area of 3D-reconstructed meshes emerged as the most predictive features for estimating TLA after applying mutual information and Boruta selection methods.

Figure 11. Model performance evaluation of XGBoost predictions. (a) Scatter plot comparing predicted and actual TLA values along the 1:1 line. (b) Residual plot showing minimal bias and random error distribution, with slight variability at higher TLA values.